\documentclass[a4paper,fleqn]{cas-dc}

\usepackage[authoryear,longnamesfirst]{natbib}
\usepackage{algorithm}
\usepackage{algpseudocode}

\newcommand{\myState}[1]{%
  \State \parbox[t]{0.7\linewidth}{\hangindent=\algorithmicindent #1}%
}



\def\tsc#1{\csdef{#1}{\textsc{\lowercase{#1}}\xspace}}
\tsc{WGM}
\tsc{QE}

\begin{document}
\let\WriteBookmarks\relax
\def\floatpagepagefraction{1}
\def\textpagefraction{.001}

\shorttitle{Black-Box Time-Series Domain Adaptation via Cross-Prompt Foundation Models}    

\shortauthors{M. T. Furqon et~al.}  

\title [mode = title]{Black-Box Time-Series Domain Adaptation via Cross-Prompt Foundation Models}  

\tnotemark[1] 

\tnotetext[1]{} 

%

\author[1]{M. T. Furqon}



\ead{muhammad_tanzil.furqon@mymail.unisa.edu.au}



\affiliation[1]{organization={STEM, University of South Australia},
            addressline={Mawson Lakes Boulevard}, 
            city={Adelaide},
            postcode={5095}, 
            country={Australia}}

\author[1]{Mahardhika Pratama}[type=editor]
\cormark[1]
\ead{dhika.pratama@unisa.edu.au}
\cortext[1]{Corresponding author}

\affiliation[2]{organization={Faculty of Electrical and Computer Engineering, University of Ljubljana},
            country={Slovenia}}

\author[2]{Igor ŠKRJANC}
\ead{Igor.Skrjanc@fe.uni-lj.si}
            







\author%
[1]
{Lin Liu}
\ead{lin.liu@unisa.edu.au}
\author[1]{Habibullah Habibullah}
\ead{habibullah.habibullah@unisa.edu.au}
\author[1]{Kutluyil Dogancay}
\ead{kutluyil.dogancay@unisa.edu.au}



\begin{abstract}
The black-box domain adaptation (BBDA) topic is developed to address the privacy and security issues where only an application programming interface (API) of the source model is available for domain adaptations. Although the BBDA topic has attracted growing research attentions, existing works mostly target the vision applications and are not directly applicable to the time-series applications possessing unique spatio-temporal characteristics. In addition, none of existing approaches have explored the strength of foundation model for black box time-series domain adaptation (BBTSDA). This paper proposes a concept of Cross-Prompt Foundation Model (CPFM) for the BBTSDA problems. CPFM is constructed under a dual branch network structure where each branch is equipped with a unique prompt to capture different characteristics of data distributions. In the domain adaptation phase, the reconstruction learning phase in the prompt and input levels is developed. All of which are built upon a time-series foundation model to overcome the spatio-temporal dynamic. Our rigorous experiments substantiate the advantage of CPFM achieving improved results with noticeable margins from its competitors in three time-series datasets of different application domains.
\end{abstract}


\begin{highlights}
\item We propose the concept of CPFM for the BBTSDA problems. It is constructed under the time-series foundation model utilizing the prompt tuning strategy. 
\item We propose the idea of dual-branch network structure where each branch implements a unique prompt and complements each other. The final output is drawn from the aggregation of each branch output.
\item We propose the idea of reconstruction learning in the prompt and input levels. The prompt reconstruction strategy creates distinct prompts generating complementary information while the input reconstruction method performs implicit domain alignment of the target domain by feeding the target domain samples without their labels. 
\item We numerically validate the advantage of CPFM using three datasets of different application domains. CPFM is capable of demonstrating the most encouraging performance outperforming SOTA algorithms with notable margins. 
\end{highlights}



\maketitle

\section{Introduction}\label{}
\begin{figure}
    \centering
    \includegraphics[width=0.9\linewidth]{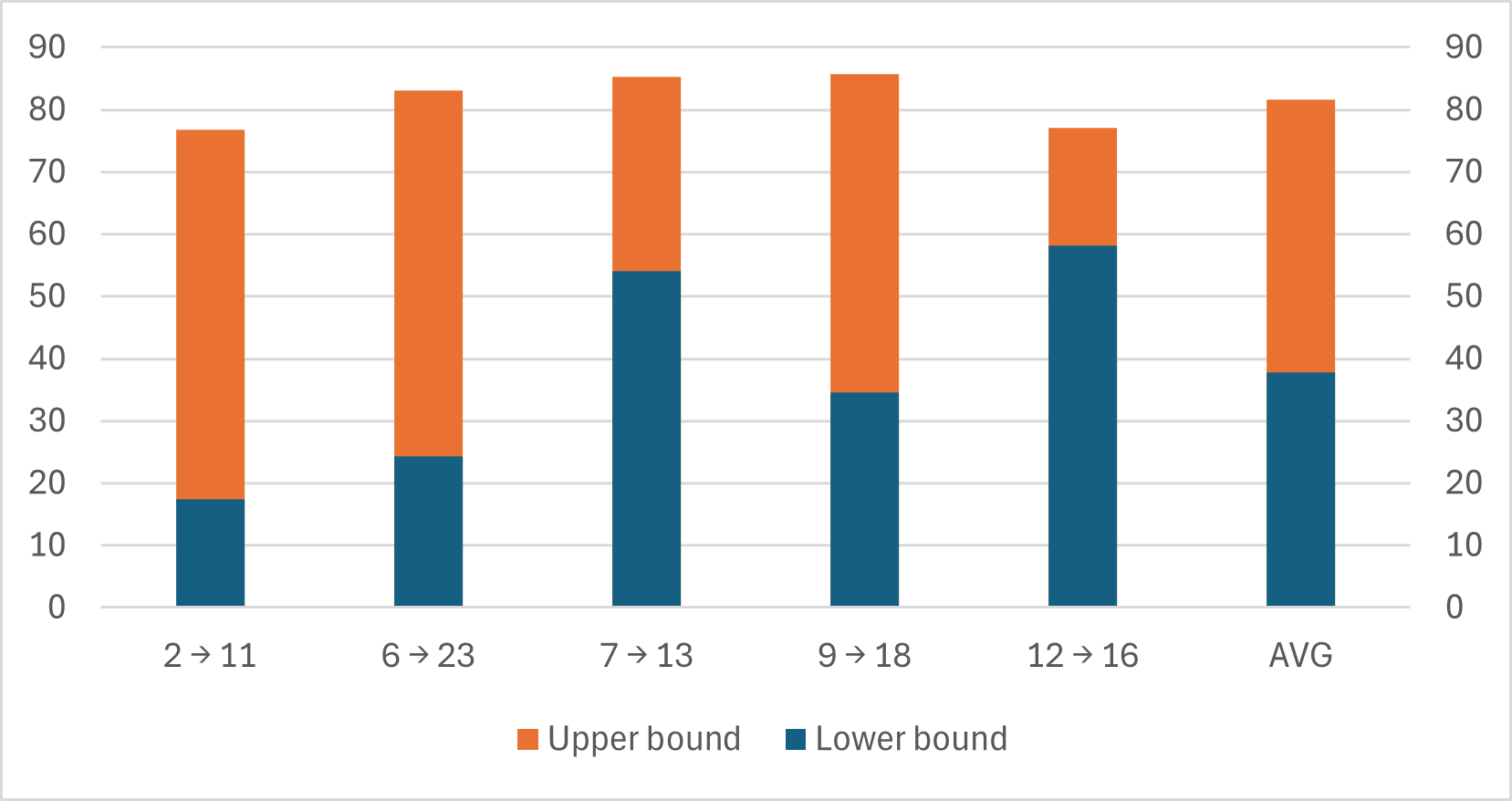}
    \caption{Lower bound vs Upper bound performance in the HAR dataset}
    \label{fig:LowUpperBound}
\end{figure}

The success of deep learning (DL) is largely attributed to the i.i.d conditions where training and testing samples follow the same distribution. However, this condition is too strict and does not mirror real-world situations where the training and deployment phases are often not the same, i.e., also known as the domain shift problem. Unsupervised domain adaptation (UDA) \cite{Ganin2014UnsupervisedDA,Kang2019ContrastiveAN,Furqon2024MixupDA} addresses this problem where the goal is to develop a model performing well on the unlabeled target domain given the labeled source domain under the presence of domain shifts between the source domain and target domain. Nonetheless, the classic UDA techniques require source-domain samples to be available, thus restricting their applications in the privacy and/or resource-constrained environments. Source-free domain adaptation (SFDA) approaches \cite{Liang2020DoWR,Karim2023CSFDAAC,Litrico2023GuidingPW,Furqon2024TimeAF} address the drawback of the traditional UDA methods accessing only the source model rather than the source-domain samples. Nevertheless, the white box nature of the SFDA approaches does not fully protect the issue of privacy. The notion of black box domain adaptation (BBDA) \cite{Liang2021DINEDA,Yang2022DivideTA} goes one step further than the SFDA where it only calls for application programming interface (API) for the target domain. Such domain adaptation is highly challenging because of the noisy pseudo label problem, i.e., the outputs of the source model obviously contain significant noise due to the domain shift problem whereas the discrepancies between two domains are not directly estimable because of the absence of source domain samples and source models. Fig \ref{fig:LowUpperBound} portrays the significant gap between the upper bound performance trained with the true class labels of the target domain and the lower bound performance delivered by the source model outputs in the human activity recognition (HAR) dataset. In a nutshell, there are many noisy pseudo labels generated by the source model. On the other hand, the BBDA problem is more challenging than the SFDA problem because of the absence of pre-trained parameters, thereby losing domain-specific information. 

The BBDA research topic has gained growing research attentions. In \cite{Liang2021DINEDA}, the so-called distill and fine-tune (DINE) is proposed to address the problem of single and multi-source black-box domain adaptation and based on the idea of two-step knowledge adaptation. BiMem is proposed in \cite{Zhang2023BlackboxUD} where it is built upon the concept of Atkinson-Shiffrin memory. BETA is designed in \cite{Yang2022DivideTA} and puts forward knowledge distillation coupled with noisy-label learning. Co-MDA is put into perspective in \cite{Liu2023CoMDAFM} where the key idea lies in the concept of multi-domain attention and the co-learning framework. In \cite{Xiao2024AdversarialEM}, the concept of dual experts is put into perspective to guide the domain adaptation process of two classifiers, while \cite{Jahan2023CurriculumGD} proposes the curriculum learning strategy to avoid noisy pseudo labels. The problem of forgotten classes in BBDA is unveiled in \cite{Zhang2024ReviewingTF}. The idea of separation and alignment (SEAL) is proposed in \cite{Liu2022SelfAlignmentFB,Xia2024ASA} where data samples are at first grouped into well-adapted and under-adapted regimes followed by the graph contrastive learning technique to improve model's representations. All these works merely focus on the vision application and are not directly applicable to the time-series domain possessing the spatio-temporal characteristic, i.e., their performance drops in the time-series problems.

To date, only two works in \cite{Ren2023SingleMultiSourceBD,Jiao2025SourceFreeBA} discuss the issue of black-box time-series domain adaptation (BBTSDA). \cite{Ren2023SingleMultiSourceBD} is framed under the teacher-student learning with the temporal consistency loss. In addition, the shapley-enhanced method is incorporated to derive the contribution of each source domain. \cite{Jiao2025SourceFreeBA} applies the knowledge distillation concept followed by local and global regularization for fault diagnosis problems. Nevertheless, the issue of noisy labels remain unsolved as the big gap to the upper bound performance still exists. Beside, none of existing approaches explore the strength of the foundation model offering performance improvements. That is, the foundation model can be shared across different nodes while maintaining domain-specific prompts to address downstream tasks. Note that, in realm of BBDA, the prompts cannot be exchanged. They are kept private to preserve the privacy issue.  

This paper proposes Cross Prompt Foundation Models (CPFM) for the BBTSDA problems. The domain adaptation stage is developed using the idea of reconstruction learning in both input and prompt level. The input reconstruction approach functions as an implicit domain adaptation of the target domain while the prompt reconstruction strategy assures distinct prompt making possible complementary information to be explored by the dual-branch network structure. The CPFM is built upon a dual-branch network structure where distinct prompts are mounted in each branch while their outputs are combined in such a way that complementary information is fused. That is, we implement the notion of prompt tuning \cite{Wang2022DualPromptCP} where the backbone model is frozen leaving only small trainable parameters, called prompts, to be tuned. All of which are built upon a time-series foundation model \cite{Goswami2024MOMENTAF} pretrained with abundant time-series datasets guaranteeing decent model's generalizations under spatio-temporal time-series problems.  Our major contributions are listed:
\begin{itemize}
    \item We propose the concept of CPFM for the BBTSDA problems. It is constructed under the time-series foundation model utilizing the prompt tuning strategy. 
    \item We propose the idea of dual-branch network structure where each branch implements a unique prompt and complements each other. The final output is drawn from the aggregation of each branch output.
    \item We propose the idea of reconstruction learning in the prompt and input levels. The prompt reconstruction strategy creates distinct prompts generating complementary information while the input reconstruction method performs implicit domain alignment of the target domain by modeling the target domain samples without their labels. 
    \item We numerically validate the advantage of CPFM using three datasets of different application domains. CPFM is capable of demonstrating the most encouraging performance outperforming SOTA algorithms with noticeable margins. 
\end{itemize}

\section{Related Works}
\subsection{Time-Series Domain Adaptation}
Time-Series Domain Adaptation (TSDA) has been studied where the goal is to overcome the temporal nature of time-series data which does not exist in the vision application \cite{Liu2023CoMDAFM} in addition to that of the domain shifts between the source domain and the target domain. There exist two approaches in this domain: adversarial-based approach and discrepancy-based approach. AdvSKM \cite{Liu2023CoMDAFM} constitutes a discrepancy-based approach using the MMD approach coupled with the spectral kernel method to minimize the domain gap between the source domain and the target domain and to take into account the temporal dependencies of the time-series samples. The association structure is designed in SASA for TSDA \cite{Cai2020TimeSD}. MDAN \cite{Furqon2024MixupDA} puts forward the idea of intermediate domain to dampen the discrepancies between the source domain and the target domain. On the other hand, the adversarial-based approach utilizes a domain discriminator to play the adversarial game reducing the domain gap. CoDATS \cite{Wilson2020MultiSourceDD} implements such concepts for TSDA in the human activity recognition problems. DAATTN combines the adversarial learning with the attention sharing mechanism \cite{Jin2021DomainAF}. SLARDA \cite{Ragab2021SelfsupervisedAD} presents an autoregressive domain discriminator for the adversarial training approach. Notwithstanding that these approaches have been successful for TSDA, they call for source-domain samples and a pretrained source model to be shared during the domain adaptation step, thus raising the privacy and storage concerns. 

\subsection{Source-Free Domain Adaptation}
The issue of privacy has led to the advent of source-free domain adaptation (SFDA) where the goal is to generalize well over the unlabeled target domain with the absence of source domain samples. Only a pretrained source model is shared for domain adaptation. \cite{Liang2020DoWR} relies on the self-training mechanism with the cluster's structure and \cite{Li2020ModelAU} utilizes the generative model to address the absence of source domain samples. \cite{Chen2022ContrastiveTA} puts forward the idea of self-supervised learning while \cite{Karim2023CSFDAAC} also uses the self-supervised learning technique combined with the notion of curriculum learning to prevent early memorization of noisy pseudo labels. The concept of loss re-weighting using the entropy estimation is put forward in \cite{Litrico2023GuidingPW}. The aforementioned methods are designed for vision applications excluding any spatio-temporal properties. The proposals of source-free time-series domain adaptation (SFTSDA) are exemplified in \cite{Zhao2023SourceFreeDA} using the GMM concept for seizure predictions and \cite{Ragab2023SourceFreeDA} integrating the time-series imputation strategy. In \cite{Furqon2024TimeAF}, the time-frequency concept is introduced for SFTSDA. Nonetheless, the SFDA concept does not fully protect the client's privacy because the source model is not shareable in many applications. That is, source-domain samples can be reconstructed by certain techniques such as the deepinversion due the presence of the pretrained source model. 

\subsection{Black-Box Domain Adaptation}
Black-Box Domain Adaptation (BBDA) goes one step ahead of SFDA where only the API of the source model is offered for domain adaptations. That is, one can only elicit soft or hard labels of the source model for further preserving the client's privacy.
\cite{Liang2021DINEDA} proposes the so-called DINE for single and multi-source BBDA using the two steps knowledge adaptation. The concept of Atkinson-Shiffrin memory is realized in \cite{Zhang2023BlackboxUD} for BBDA. The combination of multi-domain attention and co-learning is proposed in Co-MDA \cite{Liu2023CoMDAFM} for BBDA while BETA is devised in \cite{Yang2022DivideTA} using the concept of knowledge distillation and noisy-label learning. The concept of dual experts is proposed in \cite{Xiao2024AdversarialEM} while the curriculum learning approach is put forward in \cite{Jahan2023CurriculumGD}. The issue of forgotten classes in BBDA is discussed in \cite{Zhang2024ReviewingTF}. The separation and alignment (SEAL) method is put into perspective in \cite{Liu2022SelfAlignmentFB,Xia2024ASA} and achieves SOTA results in the vision applications. All these methods are designed for vision application and are not readily applicable for the time-series applications. To the best of our knowledge, the black-box time-series domain adaptation (BBTSDA) problem is only addressed in \cite{Ren2023SingleMultiSourceBD,Jiao2025SourceFreeBA} where the temporal consistency loss and the shapley-enhanced method are integrated in \cite{Ren2023SingleMultiSourceBD} while \cite{Jiao2025SourceFreeBA} presents the concept of knowledge distillation followed by local and global regularization. None of existing methods explore the advantage of foundation models possibly offering promising alternatives. That is, the foundation model can be kept fixed and shared across each node while only performing parameter efficient fine-tuning strategies for domain-specific problems. In other words, the foundation model captures general spatio-temporal characteristics of time-series data because it is pre-trained using massive time-series problems. To protect the privacy issue, the prompts providing domain-specific information can be kept private for each client. 

\section{Preliminaries}
\subsection{Problem Definition}
Given a target model $f_{\phi_t}(g_{\psi_t}(.))$ where $g_{\psi_t}(.):\mathcal{X}\rightarrow\mathcal{Z}$ is a feature extractor mapping the input space to the latent space and $f_{\phi_t}(.):\mathcal{Z}\rightarrow\mathcal{Y}$ is a projector converting the latent space to the label space, the goal of black-box time-series domain adaptation (BBTSDA) is to perform well on an unlabelled target domain $\mathcal{D}_{T}$ having $N_t$ unlabelled samples $\{x_i\}_{i=1}^{N_t}$ where $x_{i}\in\Re^{T\times D}$. $T,D$ respectively stand for the length of a time-series sample and the number of variables. The domain adaptation process is guided by $M$ black-box predictors $\{f_{\phi_{s_{i}}}(g_{\psi_{s_{i}}}(.))\}_{i=1}^{M}$ pre-trained by an $i-th$ source domain $\mathcal{D}_{S_{i}},i\in\{1,...,M\}$ consisting of $N_{s_{i}}$ labelled samples $\{(x_j,y_j)\}_{j=1}^{N_{s_{i}}}$. The underlying challenge is perceived in the issue of domain shifts where the target domain and each source domain follow an independent distribution such that $\mathcal{D}_{T}\neq\mathcal{D}_{S_{i}}\neq\mathcal{D}_{S_{j}},i,j\in\{1,...,M\}$. In addition, the BBTSDA fully preserves the client's privacy where the source-domain samples $\{(x_j,y_j)\}_{j=1}^{N_{s_{i}}}$ and the parameters of the black box predictors $\{\psi_{s_{i}},\phi_{s_{i}}\}_{i=1}^{M},i\in\{1,...,M\}$ are unavailable for domain adaptations. That is, it is only navigated by the API of the black box predictors for the target domain generating the soft-label $\hat{y}_{i}^{t}=f_{\phi_{s_i}}(g_{\psi_{s_i}}(x^{t}))$ or the hard label $\hat{C}_{i}^{t}=\arg\max_{c} \delta(f_{\phi_{s_i}}(g_{\psi_{s_i}}(x^{t})))$ where $\delta(.)$ is a softmax function. Since we exploit the foundation model in this paper, the prompts of the foundation models of the source domains are kept private. We limit our discussion in the closed-set scenario where the source domain and the target domain share the same label space $\mathcal{Y}_s=\mathcal{Y}_{t}$. 

\subsection{Foundation Model}
CPFM is built upon MOMENT \cite{Goswami2024MOMENTAF}, a family of open-source foundation models for general-purpose time-series analysis. MOMENT is pre-trained using abundant time-series datasets in the reconstruction fashion. First, a time-series is broken-down into $N$ dis-joint sub-sequences with a length of $P$ termed, patches. Each patch is projected into $D$ dimensional embedding using a trainable linear projector if unmasked or a designated learnable mask embedding if masked. These $N$ patch embedding becomes an input to the transformer model and maintains its shape $(1\times D)$. It is used to reconstruct the masked and unmasked time-series samples using a lightweight projection head. The transformer encoder follows a modification of the original transformer \cite{Raffel2019ExploringTL} where the additive bias of the layer norm is removed and placed before the residual connection. It uses relational positional embedding scheme. No decoder is applied to allow the architectural modifications for task-specific fine-tuning. 

\subsection{Prompt Tuning}
The prompt tuning concept \cite{Wang2022DualPromptCP} is implemented in the CPFM where a small-sized external parameter, namely prompt, is injected in the multi-head self-attention layer (MSA) of the foundation model leaving the backbone network frozen, thus significantly decreasing the number of trainable parameters. That is, the learning process is only localized to the prompts. The prompt tuning technique modifies the input to the MSA layer. Let $p\in\Re^{L_{p}\times D}$ be the prompt and $h_Q,h_K,h_V\in\Re^{L\times D}$ be the input query, key and value, the prompt tuning method prepends the prompts to the input token which is equivalent to concatenate the same prompt parameter to the input query, key and values \cite{Wang2022DualPromptCP} $MSA([p;h_Q],[p;h_k],[p;hv])$ where $[;]$ stands for the concatenation operation along the sequence length dimension. This leads to the increase of the output length $\Re^{(L+Lp)\times D}$.

\section{Cross Prompt Foundation Model (CPFM)}
\begin{figure*}[h]
    \centering
    \includegraphics[width=0.9\linewidth]{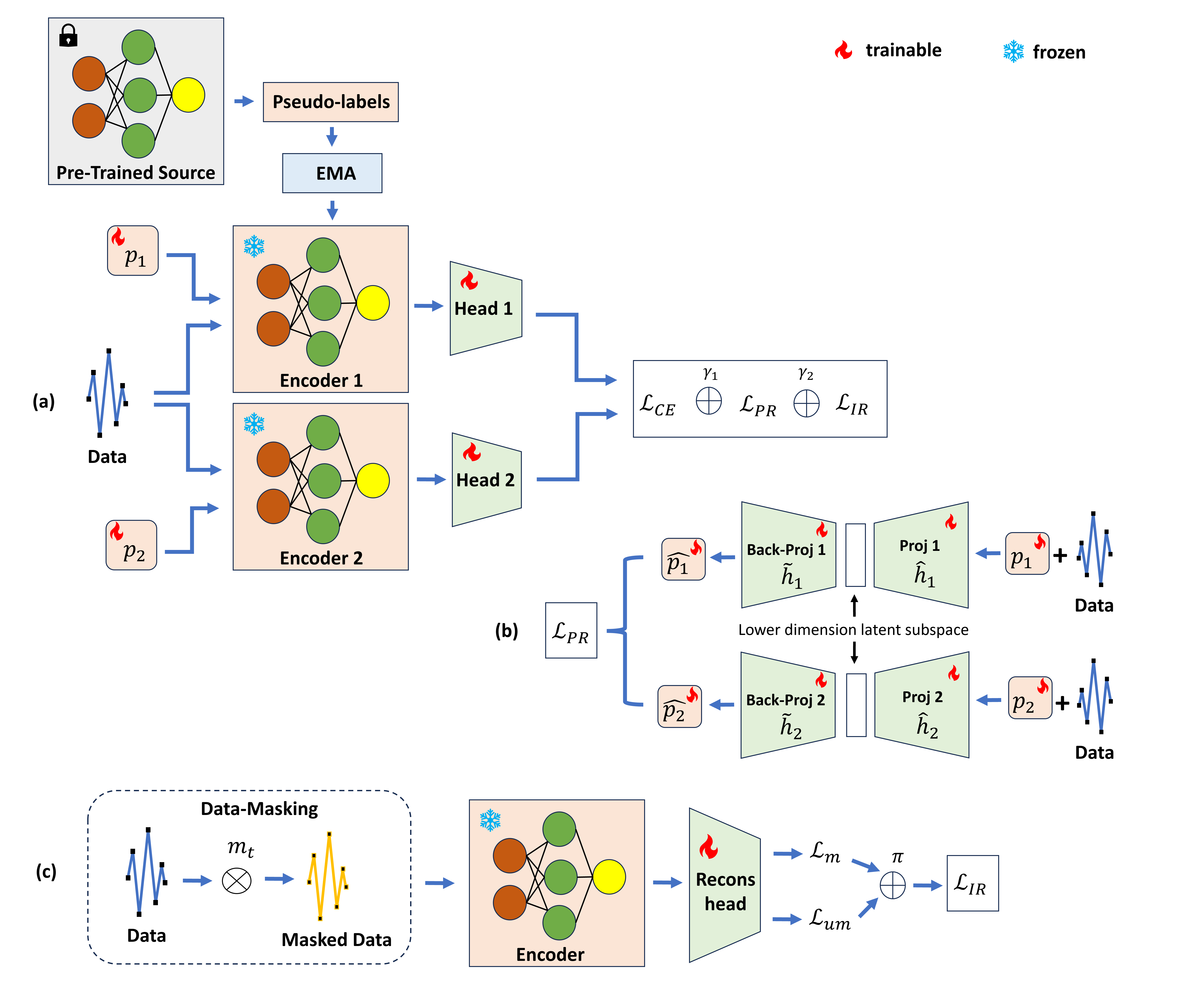}
    \caption{(a) CPFM learning policy where only the prompts and classification heads are tuned leaving the backbone network frozen; (b) Prompt Reconstruction is devised to deliver distinct prompts in each network branch; (c) Input Reconstruction learns the structure of the target domain without any label information.}
    \label{fig:cpfm}
\end{figure*}

Our algorithm, namely CPFM, is constructed under the time-series foundation model \cite{Goswami2024MOMENTAF} where the idea of prompt tuning \cite{Wang2022DualPromptCP} is integrated to adapt to downstream tasks. The dual-branch network structure is devised where each branch is inserted with unique prompts to explore different aspects of data distributions. Their outputs are aggregated to deliver final outputs and produce complementary information. The domain adaptation phase adopts the idea of reconstruction learning in the prompt level and the input level. The prompt reconstruction method is meant to generate distinct prompts offering complementary information while the input reconstruction approach performs the domain alignment step where input samples of the target domain are reconstructed. Fig. \ref{fig:cpfm} shows the workflow of our approach. 

\subsection{Dual Branch Network Structure}
CPFM is underpinned by the idea of cross-prompts working collaboratively to reject noisy pseudo-labels. This concept implements dual prompts for every source domain: $p_{i}^{1}$ and $p_{i}^{2}$ denoting respectively the first and second prompts in respect to the $i-th$ source domain while relying on the same foundation model frozen during the training process. Hence, this strategy minimizes the memory burdens because the same foundation model is applied to each source domain $\{\psi_t\}=\{\psi_{s_{i}}\},i\in\{1,...,M\}$. That is, they differ from each other only in the use of different prompts $p_{i}^{1}\neq p_{i}^{2}\neq p_{j}^{1}\neq p_{j}^{2}, i,j\{1,...,M\}, i\neq j$. In other words, their prompts are initialized differently. The embedding of the foundation model for the $i-th$ source domain can be expressed:
\begin{equation}
    g_{\psi}(x;p_{i}^{1})=MSA([p_{i}^{1};Q_{j,k}],[p_{i}^{1};K_{j,k}],[p_{i}^{1};V_{k,j}])
\end{equation}
\begin{equation}
    g_{\psi}(x;p_{i}^{2})=MSA([p_{i}^{2};Q_{j,k}],[p_{i}^{2};K_{j,k}],[p_{i}^{2};V_{k,j}])
\end{equation}
where $Q_{j,k},K_{j,k},V_{j,k}$ stand for the query, key and value of the $j-th$ head of MSA layer of the $k-th$ encoder. That is, the prompts are injected in every MSA head of the encoder layer. This mechanism results in different predictions due to the use of different linear heads fine-tuned during the training process $\{\phi_{t}^{1}\}\neq\{\phi_{t}^{2}\}\neq\{\phi_{s_{i}}\},i\in\{1,...,M\}$. Suppose that $o_{i}^{1}=\sigma(f_{\phi_{i}^{1}}(g_{\psi}(x;p_{i}^{1})))$ and $o_{i}^{2}=\sigma(f_{\phi_{i}^{2}}(g_{\psi}(x;p_{i}^{2})))$, the final output of the two networks are aggregated \cite{Wang2023FewShotCS} as follows: 
\begin{equation}\label{output}
    o_{i}=\alpha o_{i}^{1} + \beta o_{i}^{2}
\end{equation}
where $\alpha=\frac{\max_{c\in[0,C]}o_{i}^{1}}{\max_{c\in[0,C]}o_{i}^{1}+\max_{c\in[0,C]}o_{i}^{2}}$ and $\beta=\frac{\max_{c\in[0,C]}o_{i}^{2}}{\max_{c\in[0,C]}o_{i}^{1}+\max_{c\in[0,C]}o_{i}^{2}}$. $\sigma(.)$ is a softmax function. Unlike \cite{Han2018CoteachingRT,Liu2023CoMDAFM} maintaining the predictions of two networks, we only retain the aggregations of the two predictions for simplicity. \textit{Note that notwithstanding that the backbone networks are the same across the source and target domains, the prompts of the source domain are kept private and not to be shared during domain adaptations. $p_{i}^{1},p_{i}^{2}$ are the prompts of the target domain when using the $i-th$ source model as the teacher.}
\subsection{Domain Adaptation Phase}
\subsubsection{Prompt Reconstruction}
Since the prompts are high-dimensional and required to reject different types of errors in the collaborative learning mechanism, they need to be  distinct and do not contain redundant information $p_{i}^{1}\neq p_{i}^{2}$ and across all source domains. To this end, we are inspired by \cite{Chen2022MultiPromptAF} that high-dimensional data usually lies in a lower dimensional manifold and thus auto-encoders can be applied for improved alignments. The goal is to learn a domain-invariant latent subspace of denoised prompts where redundant information is removed via the reconstruction of the learned prompts. Suppose that $\hat{h}(.)$ denotes a projection function mapping the prompts $p_{i}^{1,2}$ into a lower dimensional manifold and $\tilde{h}(.)$ stands for a back-projection function projecting the vectors back into soft prompts $\hat{p}_{i}^{1,2}$. We follow the same architecture as \cite{Chen2022MultiPromptAF} where $\hat{h}(.)$ is implemented as a one-layer feed-forward network while $\tilde{h}(.)$ is a two-layer nonlinear perceptron as follows:
\begin{equation}
    \hat{h}(p_{i}^{1,2})=W_1 p_{i}^{1,2}+b_1
\end{equation}
\begin{equation}
    \tilde{h}(\hat{h})=W_3tanh(W_2\hat{h}+b_2)+b_3
\end{equation}
The objective is to minimize the prompt reconstruction loss:
\begin{equation}\label{prompt_reconstruction}
    \mathcal{L}_{PR}=\frac{1}{2M}\sum_{i=1}^{M}\sum_{j=1}^{2}||\hat{p}_{i}^{j}-p_{i}^{j}||_{2}^{2}
\end{equation}
\subsubsection{Input Reconstruction}
Given the unlabelled target domain $\mathcal{D}_{T}$, the input reconstruction process is performed to the target domain samples $x_{t}$ \cite{Furqon2024MixupDA}. That is, a binary mask $m_{t}$ is generated where $x_{t}$ is zeroed if $m_{t}=1$. The input reconstruction loss comprises masked and unmasked components as follows
\begin{equation}
    \mathcal{L}_{m}=\frac{1}{D\sum_{i=1}^{N_{t}}m_{t}}\sum_{i=1}^{N_{t}}m_{t}||\delta(f_{\phi_{t}}(g_{\psi_{t}}(x_{t})))-x_{t}||_{2}^{2}
\end{equation}
\begin{equation}
\begin{split}
    \mathcal{L}_{um}=\frac{1}{D(S-\sum_{i=1}^{N_{t}}m_{t})}\\\sum_{i=1}^{N_{t}}(1-m_{t})||\delta(f_{\phi_{t}}(g_{\psi_{t}}(x_{t})))-x_{t}||_{2}^{2}
\end{split}
\end{equation}
where $\delta(.)$ is a projector. The input reconstruction loss is defined as a combination of the masked and unmasked components. 
\begin{equation}\label{input_reconstruction}
    \mathcal{L}_{IR}=\pi\mathcal{L}_{m}+(1-\pi)\mathcal{L}_{um}
\end{equation}
where $\pi\in[0,1]$ is a trade-off constant. The input reconstruction strategy functions as an implicit domain alignment phase since it learns the structure of the target domain.  

The overall objective function is mathematically expressed:
\begin{equation}\label{totalloss}
    \mathcal{L}=\mathcal{L}_{CE}+\gamma_{1}\mathcal{L}_{PR}+\gamma_{2}\mathcal{L}_{IR}
\end{equation}
where $\gamma_{1,2}$ are a trade-off constant controlling the strength of the prompt reconstruction loss and the input reconstruction loss. $\mathcal{L}_{CE}$ stands for the cross-entropy loss computed using the pseudo-labels information or the soft label of the $i-th$ source model $\hat{y}_{i}^{t}=\sigma(f_{\phi_{s_{i}}}(g_{\psi_{s_{i}}}(x^{t};p_{i}^{s_{i}})))$ where $p_{i}^{s_{i}}$ is the corresponding prompt. Nonetheless, the soft label likely contains noises leading to poor generalizations. We apply the exponential moving average (EMA) rule to alleviate the noisy pseudo label problem \cite{Jiao2025SourceFreeBA,Xiao2024AdversarialEM}. 
\begin{equation}\label{EMA1}
    \hat{y}_{i}^{t}=\gamma\hat{y}_{i}^{t}+(1-\gamma)\hat{y}^{t}
\end{equation}
where $\hat{y}^{t}$ stands for the prediction of the target model and $\gamma$ denotes the trade-off parameters simply set to $0.7$. In other words, the prediction of the $i-th$ source model is slowly forgotten and replaced by the prediction of the target model overtime. In the first epoch, the output of the source model is computed as per the following equation to reduce inaccurate and even incomplete predictions. 
\begin{equation}    \hat{y}_{i}^{t}=\begin{cases}\hat{y}_{i}^{t},\quad\quad\quad  Top^{1}\\
    \frac{(1-\hat{y}_{i}^{t})}{K-1},\quad otherwise  
    \end{cases}
\end{equation}
where $Top^{1}$ is the top 1 index of the prediction in $\hat{y}_{i}^{t}$.
\subsection{Multi-Source Domains Case}
In realm of multi-source domain adaptation, CPFM is driven by $M$ teachers representing $M$ different but related source domains and delivering predictions of unlabelled target-domain samples. The final output is mathematically expressed as follows:
\begin{equation}
    \hat{y}^{t}=\sum_{i}^{M}\lambda_{i}\hat{y}_{i}^{t}
\end{equation}
where $\hat{y}_{i}^{t}=f_{\phi_{s_{i}}}(g_{\psi_{s_{i}}}(x^{t};p_{i}^{s_{i}}))$ and $\lambda_{i}\in[0,1]$ is a weighting coefficient of the $i-th$ source domain also known as the transferability weight. The weight determines the influence of the $i-th$ source domain toward the knowledge transfer. 

Because of the absence of source domain samples, the transferrability weight $\{\lambda_{i}\}_{i=1}^{M}$ is estimated by the difficulty of knowledge transfer reflected by the prediction's uncertainty \cite{L2024AlleviatingIP}. 
\begin{equation}
    \eta_{i}=\frac{1}{\mathcal{H}(\sigma(f_{\phi_{s_{i}}}(g_{\psi_{s_{i}}}(x^{t};p_{i}^{s_{i}}))))}=\frac{1}{\mathcal{H}(\hat{y}_{i}^{t})}
\end{equation}
where $\mathcal{H}(.)$ denotes the Shannon entropy. That is, we consider the soft outputs of the $i-th$ source model. The shannon entropy is inversely proportional to the prediction confidence where the higher the entropy the less confidence the model is. This implies similarity between the $i-th$ source domain and the target domain where a low uncertainty is seen as a high similarity between the two domains. The final transferrability weight is enumerated from the inverse of Shannon entropy. The transferrability weight is normalized. 
\begin{equation}
    \lambda_{i}=\frac{\eta_{i}}{\max_{i=1,...,M}(\eta_{i})}
\end{equation}
At each epoch, the normalized transferrability weight is updated using a moving average formula. 
\begin{equation}
    \lambda_{i}^{e}=\alpha\lambda_{i}^{e-1}+(1-\alpha)\lambda_{i}^{e}
\end{equation}
where $\alpha=\frac{N_{p}}{N_{T}}$ is a scaling coefficient, while $N_{p},N_{T}$ respectively denote the number of pseudo labels and target-domain samples.

\subsection{Algorithm}
The learning policy of CPFM is outlined in Algorithm \ref{algorithm} where, at first, the foundation models trained across diverse time-series datasets are loaded. The backbone network is frozen to enjoy generalized features of the foundation model, while the domain adaptation phase is done by tuning only the prompts. We apply the dual-branch network architecture meaning that two different prompts initialize our model and the final output is aggregated as per \eqref{output}. The source model is trained in the source domain. Once completed, they are set as the teacher models for the target domain and their outputs are obtained as per the EMA formula \eqref{EMA1} where their influence decay and is taken over by the target model as the training process runs. Once eliciting the teacher output, this knowledge is distilled to the target model by the cross entropy loss. We also calculate the input reconstruction loss as an implicit domain adaptation loss and the prompt reconstruction loss to avoid redundant prompts. Finally, the total loss is calculated as per \eqref{totalloss} and induces the parameter learning process.   

\section{Complexity Analysis}
Following the pseudo-code in Algorithm \ref{algorithm}, CPFM has several operations e.g. obtain source model soft-label, initialize or update teacher buffer, obtain target model logits, perform input reconstruction, perform prompt reconstruction, calculate model losses, and update model parameters. Suppose that ${N}$ is the total number of samples in the target dataset consisting of the number B of batches that satisfies $\sum_{b=1}^{B}N_{b}=N$, $E$ is the number of training epochs, $R$ is the size of memory buffer where $R<N$ and $M$ is the number of teacher models. Let $C$ denote the complexity of a process, the complexity of the proposed method can be written:

\begin{equation}
    \begin{split}
        C(CPFM) = C(ObtainTeacherSoftLabel)\\+C(InitializeUpdateTeacherBuffer)\\+C(ObtainTargetLogits)\\+C(InputReconstruction)\\+C(PromptReconstruction)\\+C(\mathcal{L}_{CE})+C(\mathcal{L}_{IR})+C(\mathcal{L}_{PR})\\+C(UpdateModelParameters)
    \end{split}
\end{equation}
\begin{equation}
    \begin{split}
        C(CPFM) = E.\sum_{b=1}^{B} N_{b}(O(M)+O(M.R)+O(1)\\+O(1)+O(1)+O(1)+O(1)\\+O(1)+O(1))
    \end{split}
\end{equation}
\begin{equation}
    \begin{split}
        C(CPFM) = O(E.M.\sum_{b=1}^{B} N_{b})+O(E.M.R.\sum_{b=1}^{B} N_{b})
    \end{split}
\end{equation}
since $\sum_{b=1}^{B} N_{b}=N$ and M is a small number ($M < 10$), then the complexity of CPFM can be written as:
\begin{equation}
    \begin{split}
        C(CPFM) = O(E.M.N)+O(E.M.R.N)\\
        C(CPFM) = O(E.R.N)
    \end{split}
\end{equation}
\begin{algorithm}
    \caption{CPFM}
    \label{algorithm}
    \begin{algorithmic}[1]
        \State \textbf{Input:}  Source model $f_{\phi_{s_i}}(g_{\psi}(\cdot))$, target models $f_{\phi_{t_i}^1}(g_{\psi}(\cdot))$, $f_{\phi_{t_i}^2}(g_{\psi}(\cdot))$, linear head parameters $\phi_{t}^{1}$, $\phi_{t}^{2}$, prompts $p_{i}^1$, $p_{i}^2$, prompting function $f_{\text{prompt}}$, target dataset $D_{T}=\{x_{i}\}_{i=1}^{N_T}$, number of samples $N_T$, number of epochs $E$, number of batches $B$, number of models $f$, number of source/teacher models $M$, teacher buffer $R$ 
        \State \textbf{Output:} Configuration of CPFM
        \State \textbf{Procedure:}
        \State Load the foundation model $g_{\psi}$
        \State Initialize $p_{i}^1$, $p_{i}^2$
        \State Generate prompted architecture $g_{\psi}(x_i;p_{i})$ 
              \Comment{Attach prompts to MSA layers via $f_{\text{prompt}}$}
        \State Generate $R$
        \For{$m = 1$ to $M$}
                        \myState{Forward through the teacher models $f_{\phi_{s_i}}(g_{\psi}(\cdot))$, obtain the soft-labels $\hat{y}_{i}^{t}$ as per (13)}
        \EndFor
        \For{$e = 1$ to $E$}
            \For{$b = 1$ to $B$}
                \For{$f = 1$ to 2}
                    
                        \For{$r = 1$ to $R$}
                            \myState{Initialize or buffer with $\hat{y}_{i}^{t}$ by applying EMA (Eq.\,(11)).}
                        \EndFor

                    \myState{Calculate the prompted feature by $g_{\psi}(x_i;p_{i}^f)$.}
                    \myState{Obtain the target model's logits $f_{\phi_{t_i}^f}(g_{\psi}(\cdot))$}
                    \myState{Calculate the Cross-Entropy Loss $\mathcal{L}_{CE}$.}
                    \myState{Calculate the Input Reconstruction Loss $\mathcal{L}_{IR}$ (Eq.\,(9)).}
                    \myState{Calculate the Prompt Reconstruction Loss $\mathcal{L}_{PR}$ (Eq.\,(6)).}
                    \myState{Calculate the Total Loss $\mathcal{L}$ (Eq.\,(10)).}
                    \myState{Update $\phi_{t_i}^f$ by minimizing the Total Loss $\mathcal{L}$}
                \EndFor
            \EndFor		
		\EndFor
    \end{algorithmic} 
\end{algorithm}

\begin{table}[]
\caption{Dataset characteristics (Ch: \# channels, K: \# classes, S: sample length)}\label{tab:data}
\centering
\begin{tabular}{l|ccc|cc}
\hline
\multicolumn{1}{c|}{Dataset} & Ch & K & S    & \# Training & \# Testing \\ \hline
MFD                          & 1 & 3 & 5120 & 7312             & 3604            \\
UCIHAR                       & 9 & 6 & 128  & 2300             & 990             \\
SSC                          & 1 & 5 & 3000 & 14280            & 6130            \\
 \hline
\end{tabular}
\end{table}

\section{Experiments}
\subsection{Datasets}
The advantage of our method, CPFM, is rigorously evaluated with three datasets of different application domains: human activity recognition, machine fault diagnosis and sleep stage classifications. The characteristics of the three datasets are summed up in Table \ref{tab:data} 

\noindent\textbf{HAR dataset} constitutes a human activity recognition dataset using three sensors to monitor three dimensional body movements leading to 9 channels per sample. We follow the same configuration of \cite{Ragab2023SourceFreeDA} where five cross-users experiments are set up. That is, a model is developed using a dataset of one user and subsequently evaluated with another user dataset. 

\noindent\textbf{SSC dataset} is an EEG dataset monitoring sleep stages of five different classes. We adopt the same dataset as \cite{Ragab2023SourceFreeDA} using the sleep EDF dataset. We utilize a single channel, namely Fpz-Cz and 5 cross-users experiments from 10 subjects are set up. 

\noindent\textbf{MFD dataset} is a bearing fault diagnosis problem initiated by the university of Paderborn. The fault is detected using the vibration signal and this dataset comprises four working conditions where each condition represents one domain. As with \cite{Ragab2023SourceFreeDA}, the five cross-conditions experiments are put forward to evaluate the consolidated algorithms. 

\subsection{Baseline Algorithms}
CPFM is compared with five state-of-the art black box domain adaptation methods: DINE \cite{Liang2021DINEDA}, Co-MDA \cite{Liu2023CoMDAFM}, BETA \cite{Yang2022DivideTA}, SEAL \cite{Xia2024ASA}, RFC \cite{Zhang2024ReviewingTF}. All consolidated algorithms are executed under the same computational environments, i.e., 2 NVIDIA A5000 GPU with 24 GB of RAM, using their official implementations. CPFM is developed under the pytorch library and its source code is made publicly available in \url{https://github.com/furqon3009/CPFM} All algorithms are configured under the same architecture as CPFM where the moment foundation model \cite{Goswami2024MOMENTAF} is used using the notion of prompt tuning \cite{Wang2022DualPromptCP} to ensure fair comparisons. Because of limited computational resources, we are only able to run baseline algorithms with one random seed. This is mainly due to the complexity of foundation model for our computational resources. Nevertheless, this issue should not affect the rigor of our finding because our algorithm isn't sensitive against variations of random seeds, i.e., standard deviation is small. In addition, the macro F1 score is reported rather than the accuracy because it is more accurate than accuracy in the case of class imbalance. The hyper-parameters of consolidated algorithms are selected as per their official setting. Nonetheless, the grid search is applied when their performances are surprisingly poor.  
\subsection{Numerical Results}
Table \ref{tab:har} reports the numerical results of all consolidated algorithms in the HAR dataset. CPFM (MS) denotes the average numerical results of CPFM across three random seeds while CPFM only shows the CPFM numerical results in the first seed, i.e., other algorithms are only executed under the first seed. It is clearly seen that CPFM beats other algorithm with notable margins, i.e., $11\%$ gap to DINE in the second place as shown in Table \ref{tab:har}. Other algorithms perform poorly confirming the challenge of time-series domain adaptation possessing unique spatio-temporal characteristics. Numerical results in the SSC dataset are tabulated in Table \ref{tab:eeg}. It is seen that our algorithm, CPFM, outperforms other algorithms with significant margins, i.e., over $6\%$ margin to BETA in the second place. Note that all consolidated algorithms are structured under the foundation model to ensure fair comparisons. As with other two cases, CPFM is also superior to its competitors with at least $2\%$ margin to DINE in the second place in the MFD dataset as shown in the Table \ref{tab:mfd}. Unfortunately, other algorithms don't perform well with over $10\%$ gap to CPFM and DINE. This finding confirms the advantage of CPFM over the prior arts in the black-box time-series domain adaptation. In addition, direct applications of black box domain adaptation algorithms designed for vision applications to the time-series cases are not successful. That is, the time-series domain call for special treatments to cope with their spatio-temporal natures. On the other hand, the black box domain adaptation is challenging because it relies only on the API of the source model for domain adaptation. i.e, no source data nor pretrained weights are offered for domain adaptations. The underlying rationale behind higher F1 score of CPFM than other consolidated algorithms lies in the prompt tuning strategy under the dual branch network structure coupled with the dual reconstruction learning phase in both prompt and input levels assuring distinct prompts to be learned while implicitly adapting to the structure of the target domain, i.e., the input reconstruction phase learns the underlying patterns of the target domain with the absence of any labeled samples.    
\begin{table*}[!t]
\caption{Five HAR cross-domain scenarios results in terms of MF1 score. MS means Multi-Seed}\label{tab:har}
\centering
\begin{tabular}{l|c|c|c|c|c|c}
\hline
Method        & 2 → 11           & 6 → 23           & 7 → 13             & 9 → 18              & 12 → 16              & AVG            \\ \hline
BETA          & 17.8             & 17.93            & 18.26              & 16.61               & 20.75                & 18.27          \\
CoMDA         & 4.8              & 5.26             & 5.37               & 5.56                & 4                    & 4.99           \\
DINE          & 36.92   & 33.15   & 47.47        & 23.87         & 21.26          & 32.53    \\
RFC           & 18.53            & 15.33            & 9.09               & 16.18               & 12.58                & 14.34          \\
SEAL          & 7.44             & 9.43             & 17.21              & 10.29               & 5.81                 & 10.04          \\ \hline
\textbf{CPFM} & \textbf{56.08} & \textbf{37.94} & \textbf{59.81} & \textbf{34.27} & \underline{42.66} & \textbf{46.15} \\
\textbf{CPFM (MS)} & \underline {48.60±10.89} & \underline{36.89±1.37} & \underline{56.77±11.27} & \underline{32.22±2.00} & \textbf{44.91±2.31} & \underline{44.08} \\ \hline
\end{tabular}
\end{table*}

\begin{table*}[!t]
\caption{Five SSC cross-domain scenarios results in terms of MF1 score. MS means Multi-Seed}\label{tab:eeg}
\centering
\begin{tabular}{l|c|c|c|c|c|c}
\hline
Method        & 0 → 11              & 12 → 5              & 7 → 18              & 16 → 1           & 9 → 14             & AVG            \\ \hline
BETA          & 23.88               & 22.68               & 53.97         & 49.81   & 37.67        & 37.6     \\
CoMDA         & 13.67               & 20.42               & 20.97               & 17.45            & 13.36              & 17.17          \\
DINE          & 24.85         & 24.7          & 40.96               & 37.38            & 14.88              & 28.55          \\
RFC           & 18.39               & 21.16               & 27.55               & 9.38             & 27.05              & 20.71          \\
SEAL          & 12.87               & 19.91               & 14.11               & 26.06            & 24.71              & 19.53          \\ \hline
\textbf{CPFM} & \textbf{32.17} & \textbf{35.41} & \underline{55.47} & \textbf{58.04} & \underline{37.98} & \textbf{43.81} \\
\textbf{CPFM (MS)} & \underline{31.87±0.93} & \underline{31.71±5.84} & \textbf{55.57±1.20} & \underline {56.86±3.56} & \textbf{42.97±7.06} & \underline{43.80} \\ \hline
\end{tabular}
\end{table*}

\begin{table*}[!t]
\caption{Five MFD cross-domain scenarios results in terms of MF1 score. MS means Multi-Seed}\label{tab:mfd}
\centering
\begin{tabular}{l|c|c|c|c|c|c}
\hline
Method        & 0 → 1          & 1 → 2                & 3 → 1            & 1 → 0            & 2 → 3               & AVG            \\ \hline
BETA          & \underline {20.83}    & 20.83                & 20.83            & 5.58             & 20.83               & 17.78          \\
CoMDA         & 5.56           & 5.56                 & 5.56             & 5.56             & 5.56                & 5.56           \\
DINE          & 20.81          & 32.79                & \textbf{75.17}   & \textbf{66.65}   & 32.37               & \underline{45.56} \\
RFC           & \textbf{34.67} & 35.55          & 20.83            & 20.83            & \underline{52.39}         & 32.85          \\
SEAL          & \underline {20.83}    & 20.83                & 20.83            & 5.58             & 20.83               & 17.78          \\ \hline
\textbf{CPFM} & 20.81        & \textbf{52.86} & 55.31 & \underline {51.06} & \textbf{55.57} & \textbf {47.12}    \\
\textbf{CPFM (MS)} & 20.81±0        & \underline{52.62±0.22} & \underline {55.38±0.12} & 47.32±5.74 & 51.65±4.38 & \underline{45.56}    \\ \hline
\end{tabular}
\end{table*}

\begin{table*}[!t]
\caption{Five HAR multi-source cross-domain scenarios results in terms of MF1 score. 1S for single-source, 3S and 5S for three and five sources respectively}\label{tab:harmulti}
\begin{tabular}{l|c|c|c|c|c|c}
\hline
Method    & → 11  & → 23  & → 13  & → 18  & → 16  & AVG    \\ \hline
CPFM (1S) & 56.08 & 37.94 & 59.81 & 34.27 & 42.66 & 46.15  \\
CPFM (3S) & 37.91 & 65.25 & 54.4  & 46.39 & 72.45 & 55.28  \\
CPFM (5S) & 35.78 & 72.51 & 67.39 & 51.77 & 52.42 & 55.97 \\ \hline
\end{tabular}
\end{table*}

\begin{table*}[!t]
\caption{Five SSC multi-source cross-domain scenarios results in terms of MF1 score. 1S for single-source, 3S and 5S for three and five sources respectively}\label{tab:eegmulti}
\begin{tabular}{l|c|c|c|c|c|c}
\hline
Method    & → 11  & → 5   & → 18  & → 1   & → 14  & AVG    \\ \hline
CPFM (1S) & 32.17 & 35.41 & 55.47 & 58.04 & 37.98 & 43.81  \\
CPFM (3S) & 35.41 & 41.38 & 55.33 & 49.43 & 48.54 & 46.02 \\
CPFM (5S) & 28.62 & 44.16 & 58.29 & 54.9  & 44.77 & 46.15 \\ \hline
\end{tabular}
\end{table*}

\begin{table*}[!t]
\caption{Ablation study in HAR dataset.}\label{tab:ablation}
\begin{tabular}{l|c|c|c|c|c|c}
\hline
Method                     & 2 → 11 & 6 → 23 & 7 → 13 & 9 → 18 & 12 → 16 & AVG   \\ \hline
CPFM                       & 56.08  & 37.94  & 59.81  & 34.27  & 42.66   & 46.15 \\
CPFM   (w/o Prompt)        & 47.45  & 29.58  & 37.28  & 32.83  & 41.13   & 37.65 \\
CPFM   (w/o Input Recons)  & 42.85  & 38.34  & 68.72  & 32.62  & 41.18   & 44.74 \\ 
CPFM   (w/o Prompt Recons) & 51.82  & 39.18  & 61.3   & 33.41  & 35.23   & 44.19 \\ \hline
\end{tabular}
\end{table*}

\begin{table*}[!t]
\caption{Comparisson of multi-source domain cases in HAR dataset. 3S and 5S for three and five sources respectively}\label{tab:meanmulti}
\begin{tabular}{l|c|c|c|c|c|c}
\hline
Method              & → 11  & → 23  & → 13  & → 18  & → 16  & AVG   \\ \hline
CPFM (3S)           & 37.91 & 72.44 & 65.25 & 46.39 & 54.4  & 55.28 \\
CPFM (5S)           & 35.78 & 72.51 & 67.39 & 51.77 & 52.42 & 55.97 \\ \hline
CPFM   NaiveAvg (3S) & 32.59 & 58.25 & 73.96 & 42.99 & 54.93 & 52.54 \\
CPFM   NaiveAvg (5S) & 39.86 & 57.18 & 62.32 & 52.79 & 52.74 & 52.98 \\ \hline
\end{tabular}
\end{table*}

\subsection{Multi-Source Domain Cases}
We discuss the advantage of our algorithm, CPFM, under the multi-source problems. CPFM is configured under three and five source domains respectively and tested with the HAR and SSC datasets. Numerical results of the HAR dataset are reported in Table \ref{tab:harmulti} while numerical results of the SSC dataset are tabulated in Table \ref{tab:eegmulti}. It is observed that the performance of CPFM improves steadily when using 1 source domain to 5 source domains. CPFM attains around $10\%$ improvements from 1 source to 3 sources and $11\%$ improvements from 1 source to 5 sources in realm of the HAR dataset. The same finding takes place for the SSC dataset, i.e., $3\%$ improvements from 1 source to 3 sources and $3.1\%$  improvements from 1 source to 5 sources. This result confirms the efficacy of the multi-source domain strategy of CPFM using the normalized Shannon entropy and the momentum update rule. That is, such strategy enables complementary information to be mined while mitigating detrimental impact of unrelated source domains.

\subsection{Ablation Study}
We discuss the ablation study to verify the advantage of each learning module of CPFM where our numerical results in the HAR dataset are displayed in Table \ref{tab:ablation}. We start from the efficacy of the prompt tuning strategy. That is, CPFM discards the prompt and only adjusts the classification head for domain adaptation. The absence of the prompts for domain adaptations deteriorates the performance of CPFM by about $9\%$. This finding confirms the advantage of our prompt tuning strategy for domain adaptations where it guides the representations of the foundation model to adapt to a downstream task. The advantage of the input reconstruction loss \eqref{input_reconstruction} is tested. It is perceived that CPFM loses around $2\%$ in the MF1-score without the input reconstruction mechanism. This module plays a vital role in CPFM where it models the structure of the target domain without any labels. That is, it performs an implicit domain adaptation step. We also study the effect of the prompt reconstruction mechanism \eqref{prompt_reconstruction} to CPFM. The absence of prompt reconstruction strategy brings down the performance of CPFM by over $2\%$. The prompt reconstruction strategy is crucial because it underpins the creation of distinct prompts under the dual branch network structure. Last but not least, we also investigate the performance of CPFM with the naive averaging strategy in the multi-source domain adaptation phase in which Table \ref{tab:meanmulti} exhibits our numerical results in the HAR dataset. That is, the transferability weight is set uniformly for every source domain. This modification results in significant performance drops of CPFM for both 3 and 5 source domains confirming the advantage of our normalized entropy and momentum update strategy in the multi-source domain adaptation problems.    

\subsection{TSNE Analysis}
\begin{figure}[h]
    \centering
    \includegraphics[width=0.9\linewidth]{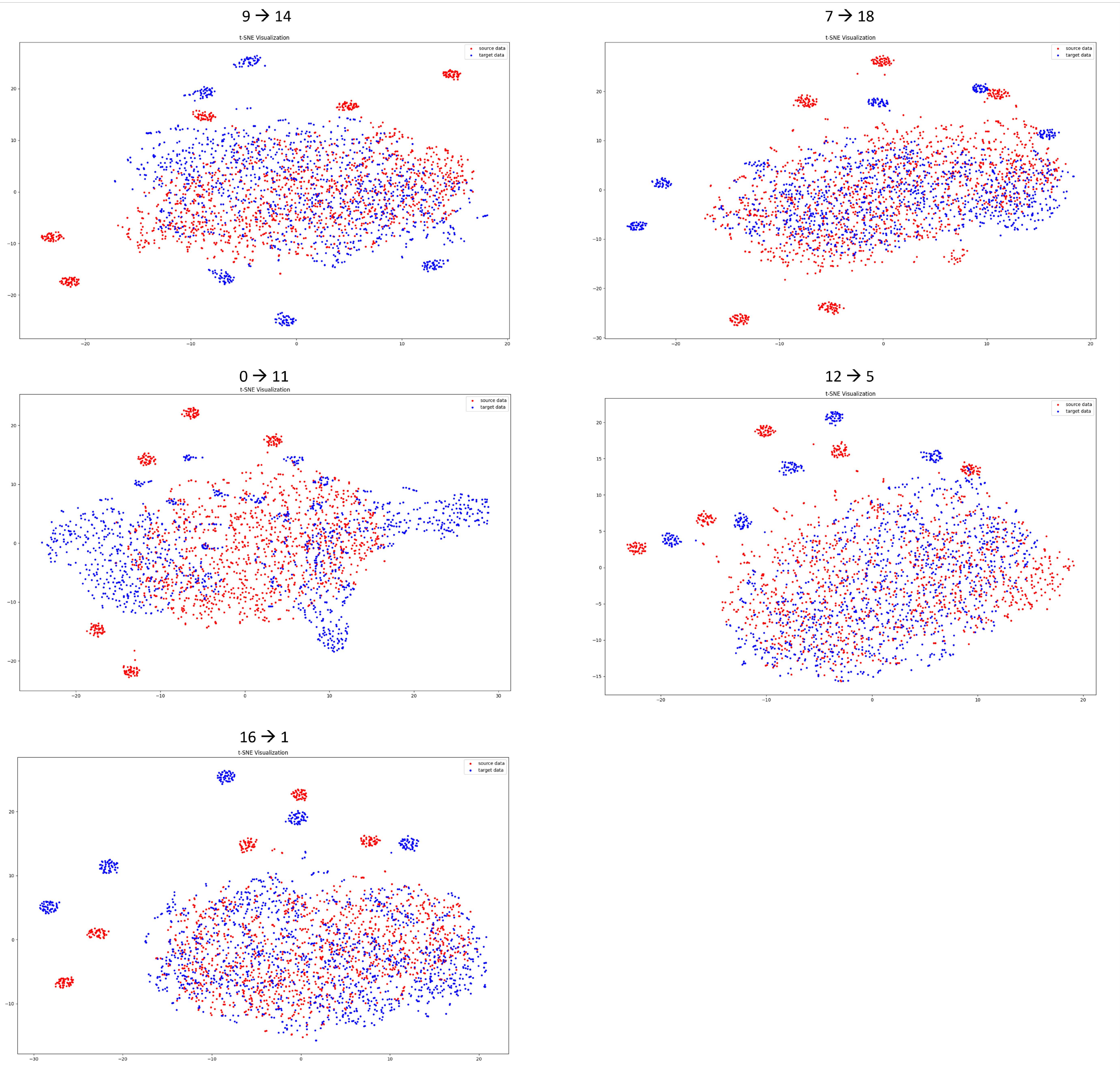}
    \caption{T-SNE of SSC Dataset by network branch 1 before training}
    \label{fig:tsneNet1Before}
\end{figure}

\begin{figure}[h]
    \centering
    \includegraphics[width=0.9\linewidth]{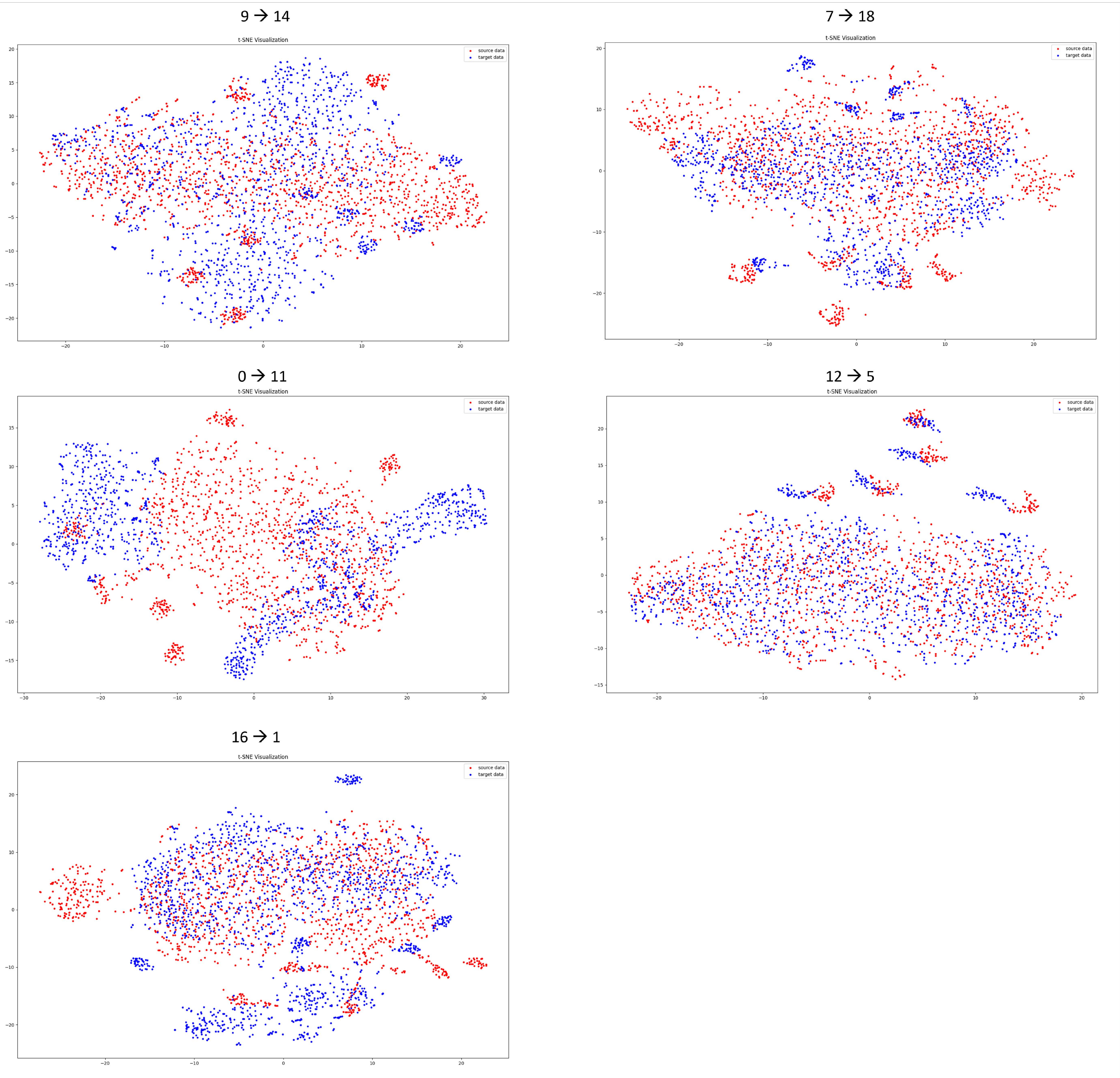}
    \caption{T-SNE of SSC Dataset by network branch 1 after training}
    \label{fig:tsneNet1After}
\end{figure}

\begin{figure}[h]
    \centering
    \includegraphics[width=0.9\linewidth]{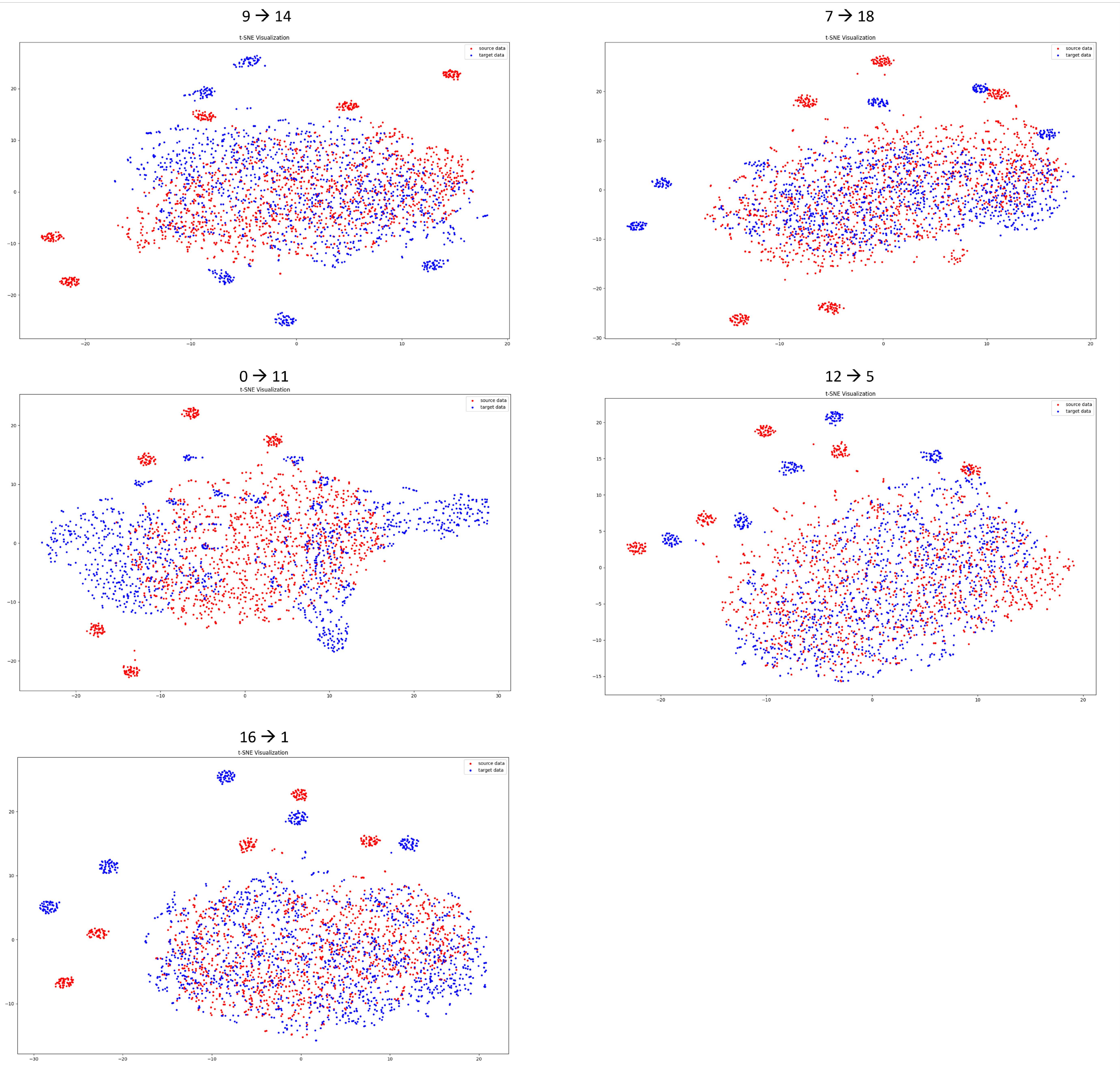}
    \caption{T-SNE of SSC Dataset by network branch 2 before training}
    \label{fig:tsneNet2Before}
\end{figure}

\begin{figure}[h]
    \centering
    \includegraphics[width=0.9\linewidth]{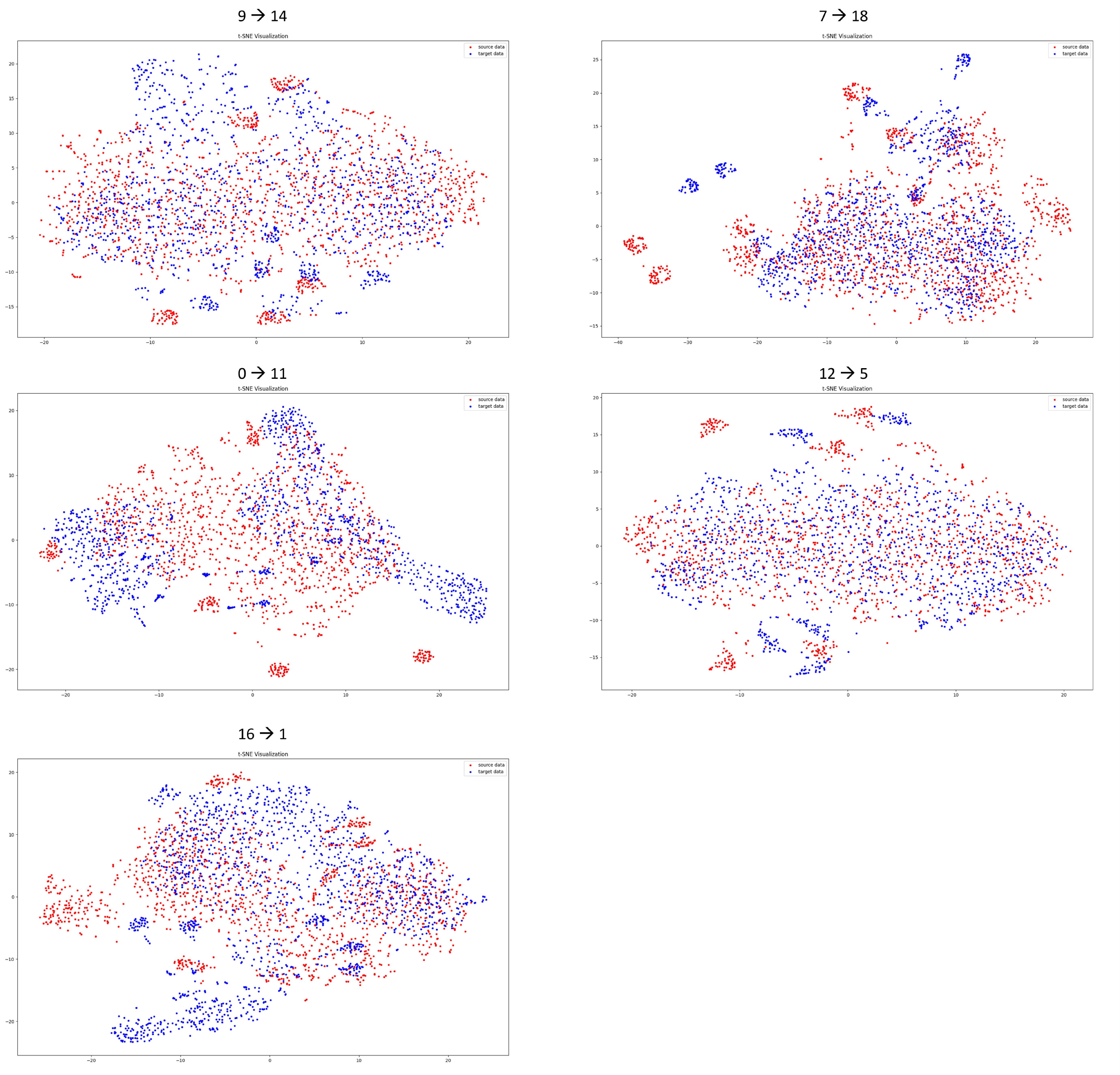}
    \caption{T-SNE of SSC Dataset by network branch 2 after training}
    \label{fig:tsneNet2After}
\end{figure}

Fig. \ref{fig:tsneNet1Before} and \ref{fig:tsneNet1After} visualize the TSNE plots of the network branch 1 before and after the training process respectively for the SSC dataset while Fig. \ref{fig:tsneNet2Before} and Fig. \ref{fig:tsneNet2After} depict the TSNE plots of the network branch 2 before and after the training process respectively for the SSC dataset. As CPFM is built upon the dual-branch network structure, the TSNE plots encompass the embedding of each network branch. It is seen that even before the training process begins, the foundation model enjoys generalizable features. These features turn to be discriminative after the training process because the prompts are learned and the backbone networks are frozen. That is, the use of adjustable prompts enhances the embedding qualities. Both network branch 1 and 2 induce the same embeddings before the training process because of the same prompts. The advantage of the prompt reconstruction strategy can be also viewed here where it generate distinct prompts inducing complementary information, i.e., the embeddings of the two branches are distinguishable after the training process, thus implying distinct prompts because the backbone network is unchanged during the training process. It is also perceived that the source and target samples are mapped closely after the training process. 

\section{Conclusion}
This paper studies the problem of black box time-series domain adaptation (BBTSDA) and proposes a novel algorithm, termed cross-prompt foundation model (CPFM) for solving the BBTSDA problem. CPFM is built upon a time-series foundation model coupled with the prompt tuning strategy. We put forward the notion of the dual-branch network structure where a unique prompt is attached to each network branch to generate complementary information. This strategy is supported with the prompt reconstruction strategy to produce distinct prompts while the input reconstruction strategy functions as the implicit domain adaptation step by modeling the target domain directly without any labels. CPFM is also adept at the multi-source domain adaptation case using the idea of normalized entropy and momentum update technique. Our numerical results confirm the advantage of CPFM over prior arts with noticeable margins in three datasets of different application domains. The performance of CPFM steadily increases with the number of source domains while the ablation study bears out the positive impact of each learning module. Our study still assumes the closed-set scenario where the source and target domains share identical label space. Our future study will be devoted to explore the category shift problem in the time-series domain adaptation.  












\clearpage
\bibliographystyle{cas-model2-names}

\bibliography{cas-refs}

\begin{thebibliography}{33}
\expandafter\ifx\csname natexlab\endcsname\relax\def\natexlab#1{#1}\fi
\providecommand{\url}[1]{\texttt{#1}}
\providecommand{\href}[2]{#2}
\providecommand{\path}[1]{#1}
\providecommand{\DOIprefix}{doi:}
\providecommand{\ArXivprefix}{arXiv:}
\providecommand{\URLprefix}{URL: }
\providecommand{\Pubmedprefix}{pmid:}
\providecommand{\doi}[1]{\href{http://dx.doi.org/#1}{\path{#1}}}
\providecommand{\Pubmed}[1]{\href{pmid:#1}{\path{#1}}}
\providecommand{\bibinfo}[2]{#2}
\ifx\xfnm\relax \def\xfnm[#1]{\unskip,\space#1}\fi
\bibitem[{Cai et~al.(2020)Cai, Chen, Li, Chen, Zhang, Ye, Li, Yang and Zhang}]{Cai2020TimeSD}
\bibinfo{author}{Cai, R.}, \bibinfo{author}{Chen, J.}, \bibinfo{author}{Li, Z.}, \bibinfo{author}{Chen, W.}, \bibinfo{author}{Zhang, K.}, \bibinfo{author}{Ye, J.}, \bibinfo{author}{Li, Z.}, \bibinfo{author}{Yang, X.}, \bibinfo{author}{Zhang, Z.}, \bibinfo{year}{2020}.
\newblock \bibinfo{title}{Time series domain adaptation via sparse associative structure alignment}, in: \bibinfo{booktitle}{AAAI Conference on Artificial Intelligence}.
\newblock \URLprefix \url{https://api.semanticscholar.org/CorpusID:229349290}.
\bibitem[{Chen et~al.(2022a)Chen, Wang, Darrell and Ebrahimi}]{Chen2022ContrastiveTA}
\bibinfo{author}{Chen, D.}, \bibinfo{author}{Wang, D.}, \bibinfo{author}{Darrell, T.}, \bibinfo{author}{Ebrahimi, S.}, \bibinfo{year}{2022}a.
\newblock \bibinfo{title}{Contrastive test-time adaptation}.
\newblock \bibinfo{journal}{2022 IEEE/CVF Conference on Computer Vision and Pattern Recognition (CVPR)} , \bibinfo{pages}{295--305}\URLprefix \url{https://api.semanticscholar.org/CorpusID:248366600}.
\bibitem[{Chen et~al.(2022b)Chen, Wu and Jiang}]{Chen2022MultiPromptAF}
\bibinfo{author}{Chen, H.}, \bibinfo{author}{Wu, Z.}, \bibinfo{author}{Jiang, Y.G.}, \bibinfo{year}{2022}b.
\newblock \bibinfo{title}{Multi-prompt alignment for multi-source unsupervised domain adaptation}.
\newblock \bibinfo{journal}{ArXiv} \bibinfo{volume}{abs/2209.15210}.
\newblock \URLprefix \url{https://api.semanticscholar.org/CorpusID:252668638}.
\bibitem[{Furqon et~al.(2024a)Furqon, Pratama, Liu, Habibullah and Doğançay}]{Furqon2024MixupDA}
\bibinfo{author}{Furqon, M.}, \bibinfo{author}{Pratama, M.}, \bibinfo{author}{Liu, L.}, \bibinfo{author}{Habibullah, H.}, \bibinfo{author}{Doğançay, K.}, \bibinfo{year}{2024}a.
\newblock \bibinfo{title}{Mixup domain adaptations for dynamic remaining useful life predictions}.
\newblock \bibinfo{journal}{Knowledge-Based Systems} \URLprefix \url{https://api.semanticscholar.org/CorpusID:269005807}.
\bibitem[{Furqon et~al.(2024b)Furqon, Pratama, Shiddiqi, Liu, Habibullah and Doğançay}]{Furqon2024TimeAF}
\bibinfo{author}{Furqon, M.T.}, \bibinfo{author}{Pratama, M.}, \bibinfo{author}{Shiddiqi, A.M.}, \bibinfo{author}{Liu, L.}, \bibinfo{author}{Habibullah, H.}, \bibinfo{author}{Doğançay, K.}, \bibinfo{year}{2024}b.
\newblock \bibinfo{title}{Time and frequency synergy for source-free time-series domain adaptations}.
\newblock \bibinfo{journal}{ArXiv} \bibinfo{volume}{abs/2410.17511}.
\newblock \URLprefix \url{https://api.semanticscholar.org/CorpusID:273532619}.
\bibitem[{Ganin and Lempitsky(2014)}]{Ganin2014UnsupervisedDA}
\bibinfo{author}{Ganin, Y.}, \bibinfo{author}{Lempitsky, V.S.}, \bibinfo{year}{2014}.
\newblock \bibinfo{title}{Unsupervised domain adaptation by backpropagation}, in: \bibinfo{booktitle}{International Conference on Machine Learning}.
\newblock \URLprefix \url{https://api.semanticscholar.org/CorpusID:6755881}.
\bibitem[{Goswami et~al.(2024)Goswami, Szafer, Choudhry, Cai, Li and Dubrawski}]{Goswami2024MOMENTAF}
\bibinfo{author}{Goswami, M.}, \bibinfo{author}{Szafer, K.}, \bibinfo{author}{Choudhry, A.}, \bibinfo{author}{Cai, Y.}, \bibinfo{author}{Li, S.}, \bibinfo{author}{Dubrawski, A.}, \bibinfo{year}{2024}.
\newblock \bibinfo{title}{Moment: A family of open time-series foundation models}.
\newblock \bibinfo{journal}{ArXiv} \bibinfo{volume}{abs/2402.03885}.
\newblock \URLprefix \url{https://api.semanticscholar.org/CorpusID:267500205}.
\bibitem[{Han et~al.(2018)Han, Yao, Yu, Niu, Xu, Hu, Tsang and Sugiyama}]{Han2018CoteachingRT}
\bibinfo{author}{Han, B.}, \bibinfo{author}{Yao, Q.}, \bibinfo{author}{Yu, X.}, \bibinfo{author}{Niu, G.}, \bibinfo{author}{Xu, M.}, \bibinfo{author}{Hu, W.}, \bibinfo{author}{Tsang, I.W.H.}, \bibinfo{author}{Sugiyama, M.}, \bibinfo{year}{2018}.
\newblock \bibinfo{title}{Co-teaching: Robust training of deep neural networks with extremely noisy labels}, in: \bibinfo{booktitle}{Neural Information Processing Systems}.
\newblock \URLprefix \url{https://api.semanticscholar.org/CorpusID:52065462}.
\bibitem[{Jahan and Savakis(2023)}]{Jahan2023CurriculumGD}
\bibinfo{author}{Jahan, C.S.}, \bibinfo{author}{Savakis, A.E.}, \bibinfo{year}{2023}.
\newblock \bibinfo{title}{Curriculum guided domain adaptation in the dark}.
\newblock \bibinfo{journal}{IEEE Transactions on Artificial Intelligence} \bibinfo{volume}{5}, \bibinfo{pages}{2604--2614}.
\newblock \URLprefix \url{https://api.semanticscholar.org/CorpusID:260378920}.
\bibitem[{Jiao et~al.(2025)Jiao, Zhang, Li, Liu and Lin}]{Jiao2025SourceFreeBA}
\bibinfo{author}{Jiao, J.}, \bibinfo{author}{Zhang, T.}, \bibinfo{author}{Li, H.}, \bibinfo{author}{Liu, H.}, \bibinfo{author}{Lin, J.}, \bibinfo{year}{2025}.
\newblock \bibinfo{title}{Source-free black-box adaptation for machine fault diagnosis}.
\newblock \bibinfo{journal}{IEEE Transactions on Industrial Informatics} \URLprefix \url{https://api.semanticscholar.org/CorpusID:275874232}.
\bibitem[{Jin et~al.(2021)Jin, Park, Maddix, Wang and Yan}]{Jin2021DomainAF}
\bibinfo{author}{Jin, X.}, \bibinfo{author}{Park, Y.}, \bibinfo{author}{Maddix, D.C.}, \bibinfo{author}{Wang, B.}, \bibinfo{author}{Yan, X.}, \bibinfo{year}{2021}.
\newblock \bibinfo{title}{Domain adaptation for time series forecasting via attention sharing}, in: \bibinfo{booktitle}{International Conference on Machine Learning}.
\newblock \URLprefix \url{https://api.semanticscholar.org/CorpusID:235421595}.
\bibitem[{Kang et~al.(2019)Kang, Jiang, Yang and Hauptmann}]{Kang2019ContrastiveAN}
\bibinfo{author}{Kang, G.}, \bibinfo{author}{Jiang, L.}, \bibinfo{author}{Yang, Y.}, \bibinfo{author}{Hauptmann, A.}, \bibinfo{year}{2019}.
\newblock \bibinfo{title}{Contrastive adaptation network for unsupervised domain adaptation}.
\newblock \bibinfo{journal}{2019 IEEE/CVF Conference on Computer Vision and Pattern Recognition (CVPR)} , \bibinfo{pages}{4888--4897}\URLprefix \url{https://api.semanticscholar.org/CorpusID:57572938}.
\bibitem[{Karim et~al.(2023)Karim, Mithun, Rajvanshi, Chiu, Samarasekera and Rahnavard}]{Karim2023CSFDAAC}
\bibinfo{author}{Karim, N.}, \bibinfo{author}{Mithun, N.C.}, \bibinfo{author}{Rajvanshi, A.}, \bibinfo{author}{Chiu, H.P.}, \bibinfo{author}{Samarasekera, S.}, \bibinfo{author}{Rahnavard, N.}, \bibinfo{year}{2023}.
\newblock \bibinfo{title}{C-sfda: A curriculum learning aided self-training framework for efficient source free domain adaptation}.
\newblock \bibinfo{journal}{ArXiv} \bibinfo{volume}{abs/2303.17132}.
\newblock \URLprefix \url{https://api.semanticscholar.org/CorpusID:257833516}.
\bibitem[{Li et~al.(2020)Li, Jiao, Cao, Wong and Wu}]{Li2020ModelAU}
\bibinfo{author}{Li, R.}, \bibinfo{author}{Jiao, Q.}, \bibinfo{author}{Cao, W.}, \bibinfo{author}{Wong, H.S.}, \bibinfo{author}{Wu, S.}, \bibinfo{year}{2020}.
\newblock \bibinfo{title}{Model adaptation: Unsupervised domain adaptation without source data}.
\newblock \bibinfo{journal}{2020 IEEE/CVF Conference on Computer Vision and Pattern Recognition (CVPR)} , \bibinfo{pages}{9638--9647}\URLprefix \url{https://api.semanticscholar.org/CorpusID:219979590}.
\bibitem[{Liang et~al.(2020)Liang, Hu and Feng}]{Liang2020DoWR}
\bibinfo{author}{Liang, J.}, \bibinfo{author}{Hu, D.}, \bibinfo{author}{Feng, J.}, \bibinfo{year}{2020}.
\newblock \bibinfo{title}{Do we really need to access the source data? source hypothesis transfer for unsupervised domain adaptation}, in: \bibinfo{booktitle}{International Conference on Machine Learning}.
\newblock \URLprefix \url{https://api.semanticscholar.org/CorpusID:211205159}.
\bibitem[{Liang et~al.(2021)Liang, Hu, Feng and He}]{Liang2021DINEDA}
\bibinfo{author}{Liang, J.}, \bibinfo{author}{Hu, D.}, \bibinfo{author}{Feng, J.}, \bibinfo{author}{He, R.}, \bibinfo{year}{2021}.
\newblock \bibinfo{title}{Dine: Domain adaptation from single and multiple black-box predictors}.
\newblock \bibinfo{journal}{2022 IEEE/CVF Conference on Computer Vision and Pattern Recognition (CVPR)} , \bibinfo{pages}{7993--8003}\URLprefix \url{https://api.semanticscholar.org/CorpusID:244800743}.
\bibitem[{Litrico et~al.(2023)Litrico, Bue and Morerio}]{Litrico2023GuidingPW}
\bibinfo{author}{Litrico, M.}, \bibinfo{author}{Bue, A.D.}, \bibinfo{author}{Morerio, P.}, \bibinfo{year}{2023}.
\newblock \bibinfo{title}{Guiding pseudo-labels with uncertainty estimation for source-free unsupervised domain adaptation}.
\newblock \bibinfo{journal}{ArXiv} \bibinfo{volume}{abs/2303.03770}.
\newblock \URLprefix \url{https://api.semanticscholar.org/CorpusID:257378054}.
\bibitem[{Liu et~al.(2022)Liu, Zhou, Ye and Li}]{Liu2022SelfAlignmentFB}
\bibinfo{author}{Liu, C.}, \bibinfo{author}{Zhou, L.}, \bibinfo{author}{Ye, M.}, \bibinfo{author}{Li, X.}, \bibinfo{year}{2022}.
\newblock \bibinfo{title}{Self-alignment for black-box domain adaptation of image classification}.
\newblock \bibinfo{journal}{IEEE Signal Processing Letters} \bibinfo{volume}{29}, \bibinfo{pages}{1709--1713}.
\newblock \URLprefix \url{https://api.semanticscholar.org/CorpusID:251473244}.
\bibitem[{Liu et~al.(2023)Liu, Xi, Li, Xu, Bai and Zhao}]{Liu2023CoMDAFM}
\bibinfo{author}{Liu, X.}, \bibinfo{author}{Xi, W.}, \bibinfo{author}{Li, W.}, \bibinfo{author}{Xu, D.}, \bibinfo{author}{Bai, G.}, \bibinfo{author}{Zhao, J.}, \bibinfo{year}{2023}.
\newblock \bibinfo{title}{Co-mda: Federated multisource domain adaptation on black-box models}.
\newblock \bibinfo{journal}{IEEE Transactions on Circuits and Systems for Video Technology} \bibinfo{volume}{33}, \bibinfo{pages}{7658--7670}.
\newblock \URLprefix \url{https://api.semanticscholar.org/CorpusID:258776491}.
\bibitem[{L{\"u} et~al.(2024)L{\"u}, Kang and Li}]{L2024AlleviatingIP}
\bibinfo{author}{L{\"u}, S.}, \bibinfo{author}{Kang, M.}, \bibinfo{author}{Li, X.}, \bibinfo{year}{2024}.
\newblock \bibinfo{title}{Alleviating imbalanced pseudo-label distribution: Self-supervised multi-source domain adaptation with label-specific confidence}.
\newblock \bibinfo{journal}{Proceedings of the Thirty-ThirdInternational Joint Conference on Artificial Intelligence} \URLprefix \url{https://api.semanticscholar.org/CorpusID:271508637}.
\bibitem[{Raffel et~al.(2019)Raffel, Shazeer, Roberts, Lee, Narang, Matena, Zhou, Li and Liu}]{Raffel2019ExploringTL}
\bibinfo{author}{Raffel, C.}, \bibinfo{author}{Shazeer, N.M.}, \bibinfo{author}{Roberts, A.}, \bibinfo{author}{Lee, K.}, \bibinfo{author}{Narang, S.}, \bibinfo{author}{Matena, M.}, \bibinfo{author}{Zhou, Y.}, \bibinfo{author}{Li, W.}, \bibinfo{author}{Liu, P.J.}, \bibinfo{year}{2019}.
\newblock \bibinfo{title}{Exploring the limits of transfer learning with a unified text-to-text transformer}.
\newblock \bibinfo{journal}{J. Mach. Learn. Res.} \bibinfo{volume}{21}, \bibinfo{pages}{140:1--140:67}.
\newblock \URLprefix \url{https://api.semanticscholar.org/CorpusID:204838007}.
\bibitem[{Ragab et~al.(2021)Ragab, Eldele, Chen, Wu, Kwoh and Li}]{Ragab2021SelfsupervisedAD}
\bibinfo{author}{Ragab, M.}, \bibinfo{author}{Eldele, E.}, \bibinfo{author}{Chen, Z.}, \bibinfo{author}{Wu, M.}, \bibinfo{author}{Kwoh, C.}, \bibinfo{author}{Li, X.}, \bibinfo{year}{2021}.
\newblock \bibinfo{title}{Self-supervised autoregressive domain adaptation for time series data}.
\newblock \bibinfo{journal}{IEEE transactions on neural networks and learning systems} \bibinfo{volume}{PP}.
\newblock \URLprefix \url{https://api.semanticscholar.org/CorpusID:244729392}.
\bibitem[{Ragab et~al.(2023)Ragab, Eldele, Wu, Foo, Li and Chen}]{Ragab2023SourceFreeDA}
\bibinfo{author}{Ragab, M.}, \bibinfo{author}{Eldele, E.}, \bibinfo{author}{Wu, M.}, \bibinfo{author}{Foo, C.S.}, \bibinfo{author}{Li, X.}, \bibinfo{author}{Chen, Z.}, \bibinfo{year}{2023}.
\newblock \bibinfo{title}{Source-free domain adaptation with temporal imputation for time series data}.
\newblock \bibinfo{journal}{Proceedings of the 29th ACM SIGKDD Conference on Knowledge Discovery and Data Mining} \URLprefix \url{https://api.semanticscholar.org/CorpusID:259937854}.
\bibitem[{Ren and Cheng(2023)}]{Ren2023SingleMultiSourceBD}
\bibinfo{author}{Ren, L.}, \bibinfo{author}{Cheng, X.}, \bibinfo{year}{2023}.
\newblock \bibinfo{title}{Single/multi-source black-box domain adaption for sensor time series data.}
\newblock \bibinfo{journal}{IEEE transactions on cybernetics} \bibinfo{volume}{PP}.
\newblock \URLprefix \url{https://api.semanticscholar.org/CorpusID:261073969}.
\bibitem[{Wang et~al.(2023)Wang, Yang, Tan, Bai and Zhou}]{Wang2023FewShotCS}
\bibinfo{author}{Wang, L.}, \bibinfo{author}{Yang, X.}, \bibinfo{author}{Tan, H.}, \bibinfo{author}{Bai, X.}, \bibinfo{author}{Zhou, F.}, \bibinfo{year}{2023}.
\newblock \bibinfo{title}{Few-shot class-incremental sar target recognition based on hierarchical embedding and incremental evolutionary network}.
\newblock \bibinfo{journal}{IEEE Transactions on Geoscience and Remote Sensing} \bibinfo{volume}{61}, \bibinfo{pages}{1--11}.
\newblock \URLprefix \url{https://api.semanticscholar.org/CorpusID:257147398}.
\bibitem[{Wang et~al.(2022)Wang, Zhang, Ebrahimi, Sun, Zhang, Lee, Ren, Su, Perot, Dy and Pfister}]{Wang2022DualPromptCP}
\bibinfo{author}{Wang, Z.}, \bibinfo{author}{Zhang, Z.}, \bibinfo{author}{Ebrahimi, S.}, \bibinfo{author}{Sun, R.}, \bibinfo{author}{Zhang, H.}, \bibinfo{author}{Lee, C.Y.}, \bibinfo{author}{Ren, X.}, \bibinfo{author}{Su, G.}, \bibinfo{author}{Perot, V.}, \bibinfo{author}{Dy, J.G.}, \bibinfo{author}{Pfister, T.}, \bibinfo{year}{2022}.
\newblock \bibinfo{title}{Dualprompt: Complementary prompting for rehearsal-free continual learning}.
\newblock \bibinfo{journal}{ArXiv} \bibinfo{volume}{abs/2204.04799}.
\newblock \URLprefix \url{https://api.semanticscholar.org/CorpusID:248085201}.
\bibitem[{Wilson et~al.(2020)Wilson, Doppa and Cook}]{Wilson2020MultiSourceDD}
\bibinfo{author}{Wilson, G.}, \bibinfo{author}{Doppa, J.R.}, \bibinfo{author}{Cook, D.J.}, \bibinfo{year}{2020}.
\newblock \bibinfo{title}{Multi-source deep domain adaptation with weak supervision for time-series sensor data}.
\newblock \bibinfo{journal}{Proceedings of the 26th ACM SIGKDD International Conference on Knowledge Discovery \& Data Mining} \URLprefix \url{https://api.semanticscholar.org/CorpusID:218863107}.
\bibitem[{Xia et~al.(2024)Xia, Zhao, Lyu, Huang, Hu, Chen and Wang}]{Xia2024ASA}
\bibinfo{author}{Xia, M.}, \bibinfo{author}{Zhao, J.}, \bibinfo{author}{Lyu, G.}, \bibinfo{author}{Huang, Z.}, \bibinfo{author}{Hu, T.}, \bibinfo{author}{Chen, G.}, \bibinfo{author}{Wang, H.}, \bibinfo{year}{2024}.
\newblock \bibinfo{title}{A separation and alignment framework for black-box domain adaptation}, in: \bibinfo{booktitle}{AAAI Conference on Artificial Intelligence}.
\newblock \URLprefix \url{https://api.semanticscholar.org/CorpusID:268708817}.
\bibitem[{Xiao et~al.(2024)Xiao, Ye, He, Li, Tang and Zhu}]{Xiao2024AdversarialEM}
\bibinfo{author}{Xiao, S.}, \bibinfo{author}{Ye, M.}, \bibinfo{author}{He, Q.}, \bibinfo{author}{Li, S.}, \bibinfo{author}{Tang, S.}, \bibinfo{author}{Zhu, X.}, \bibinfo{year}{2024}.
\newblock \bibinfo{title}{Adversarial experts model for black-box domain adaptation}, in: \bibinfo{booktitle}{ACM Multimedia}.
\newblock \URLprefix \url{https://api.semanticscholar.org/CorpusID:273645973}.
\bibitem[{Yang et~al.(2022)Yang, Peng, Wang, Zhu, Feng, Xie and You}]{Yang2022DivideTA}
\bibinfo{author}{Yang, J.}, \bibinfo{author}{Peng, X.}, \bibinfo{author}{Wang, K.}, \bibinfo{author}{Zhu, Z.H.}, \bibinfo{author}{Feng, J.}, \bibinfo{author}{Xie, L.}, \bibinfo{author}{You, Y.}, \bibinfo{year}{2022}.
\newblock \bibinfo{title}{Divide to adapt: Mitigating confirmation bias for domain adaptation of black-box predictors}.
\newblock \bibinfo{journal}{ArXiv} \bibinfo{volume}{abs/2205.14467}.
\newblock \URLprefix \url{https://api.semanticscholar.org/CorpusID:249191952}.
\bibitem[{Zhang et~al.(2023)Zhang, Huang, Jiang and Lu}]{Zhang2023BlackboxUD}
\bibinfo{author}{Zhang, J.}, \bibinfo{author}{Huang, J.}, \bibinfo{author}{Jiang, X.Q.}, \bibinfo{author}{Lu, S.}, \bibinfo{year}{2023}.
\newblock \bibinfo{title}{Black-box unsupervised domain adaptation with bi-directional atkinson-shiffrin memory}.
\newblock \bibinfo{journal}{2023 IEEE/CVF International Conference on Computer Vision (ICCV)} , \bibinfo{pages}{11737--11748}\URLprefix \url{https://api.semanticscholar.org/CorpusID:261214533}.
\bibitem[{Zhang et~al.(2024)Zhang, Shen, L{\"u} and Zhang}]{Zhang2024ReviewingTF}
\bibinfo{author}{Zhang, S.}, \bibinfo{author}{Shen, C.}, \bibinfo{author}{L{\"u}, S.}, \bibinfo{author}{Zhang, Z.}, \bibinfo{year}{2024}.
\newblock \bibinfo{title}{Reviewing the forgotten classes for domain adaptation of black-box predictors}, in: \bibinfo{booktitle}{AAAI Conference on Artificial Intelligence}.
\newblock \URLprefix \url{https://api.semanticscholar.org/CorpusID:268696714}.
\bibitem[{Zhao et~al.(2023)Zhao, Feng, Li, Song, Liang and Chen}]{Zhao2023SourceFreeDA}
\bibinfo{author}{Zhao, Y.}, \bibinfo{author}{Feng, S.}, \bibinfo{author}{Li, C.}, \bibinfo{author}{Song, R.}, \bibinfo{author}{Liang, D.}, \bibinfo{author}{Chen, X.}, \bibinfo{year}{2023}.
\newblock \bibinfo{title}{Source-free domain adaptation for privacy-preserving seizure prediction}.
\newblock \bibinfo{journal}{IEEE Transactions on Industrial Informatics} \URLprefix \url{https://api.semanticscholar.org/CorpusID:260412519}.

\end{thebibliography}



\end{document}